\documentclass[runningheads]{llncs}

\usepackage{eccv}

\usepackage{eccvabbrv}

\usepackage{graphicx}
\usepackage{booktabs}

\usepackage[accsupp]{axessibility}  

\usepackage{hyperref}

\usepackage{orcidlink}

\begin{document}

\title{Revisiting Scene Graph Generation from the Perspective of Detector-Conditioned Reachability} 
 \titlerunning{Dual-SGG}

\author{Runfeng Qu\inst{1,5}\orcidlink{0009-0008-7885-8812} \and
Pia K Bideau\inst{3}\orcidlink{0000-0001-8145-1732} \and
Ole Hall\inst{1,5}\orcidlink{0009-0001-7595-7195} \and
Julie Ouerfelli-Ethier\inst{2,5}\orcidlink{0000-0003-4886-3537} \and
Klaus Obermayer\inst{1,4,5}\orcidlink{0000-0002-5057-6142} \and
Olaf Hellwich\inst{1,5}\orcidlink{0000-0002-2871-9266}
}

\authorrunning{R.~Qu et al.}

\institute{Technische Universität Berlin, Germany \and
Humboldt Universität zu Berlin, Germany \and
Univ. Grenoble Alpes, Inria, CNRS, Grenoble INP, LJK, France \and
Bernstein Center for Computational Neuroscience, Germany \and
Science of Intelligence Research Cluster of Excellence, Germany\\
\email{runfeng.qu@campus.tu-berlin.de}}

\maketitle

\begin{abstract}
Scene graph generation (SGG) approaches can be broadly classified into detector-based and query-based methods according to their underlying reasoning mechanisms. However, the discrepancy in their predictive behaviors, induced by these distinct mechanisms, has not been systematically analyzed. In this work, we design a controlled experimental setup to examine prediction discrepancies from the perspective of detector-conditioned reachability. The results suggest clear complementary clues. Motivated by this observation, we introduce a Dual-SGG method that consolidates both reasoning mechanisms via a dual-query design, thereby leveraging the complementary predictive behaviors of both detector-based and query-based methods. Extensive experiments on the Visual Genome, Open Images v6, and GQA-200 datasets demonstrate the effectiveness of the proposed method. Code is available at: \href{https://github.com/runfeng-q/Revisiting-Scene-Graph-Generation-from-the-Perspective-of-Detector-Conditioned-Reachability}{Dual-SGG}.
\end{abstract}
\section{Introduction}
\label{sec: intro}
Scene graphs represent images as graphs, with nodes as entities and edges as pairwise relations. Each relationship is typically expressed as a (subject, predicate, object) triplet. Because of their high-level image understanding, scene graphs have been widely used in image captioning \cite{Nguyen_2021_ICCV, xu2019scene}, visual question answering \cite{shi2019explainable}, and image generation \cite{johnson2018image, li2019pastegan}.

Existing scene graph generation (SGG) approaches can be broadly categorized into \textbf{detector-based} and \textbf{query-based} methods according to their underlying reasoning mechanisms. Detector-based models employ an internal object detector to localize entities and subsequently perform reasoning over pairs of detected entities to predict triplets. Conceptually, these models exhibit difficulties in predicting triplets whose subject or object instances are semantically or spatially uncovered by detections from the employed object detector. Here, we term this problem the \textbf{detector constraint}. To eliminate this detector constraint, query-based models introduce learnable triplet queries, inspired by the detection transformer (DETR) \cite{carion2020end} paradigm, thereby enabling end-to-end triplet prediction without the need to enumerate entity pairs explicitly.  In current studies, SGG models are predominantly evaluated using aggregate metrics such as Recall and mean-Recall, which quantify overall triplet prediction performance. However, these aggregate metrics do not elucidate the distinct predictive behaviors induced by the differing reasoning mechanisms. In particular, it remains unclear whether query-based models offer improvements on triplets that are challenging for detector-based models to detect due to the detector constraint.
\begin{figure}[tb]
  \centering
  \begin{subfigure}[b]{0.15\linewidth}
    \centering
    \includegraphics[width=\linewidth]{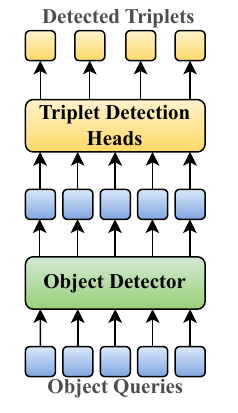}
    \caption{}
    \label{fig: Detector-based SGG}
  \end{subfigure}\hfill
  \begin{subfigure}[b]{0.15\linewidth}
    \centering
    \includegraphics[width=\linewidth]{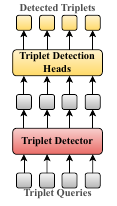}
    \caption{}
    \label{fig: Queries-based SGG}
  \end{subfigure}\hfill
  \begin{subfigure}[b]{0.3\linewidth}
    \centering
    \includegraphics[width=\linewidth]{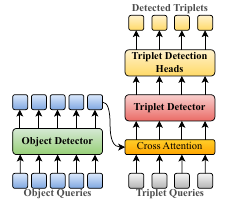}
    \caption{}
    \label{fig: Queries-based SGG with prior knowledge}
  \end{subfigure}\hfill
  \begin{subfigure}[b]{0.3\linewidth}
    \centering
    \includegraphics[width=\linewidth]{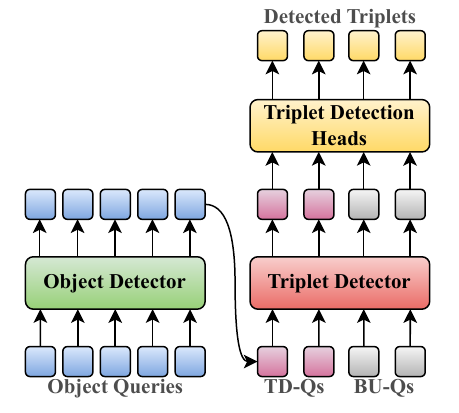}
    \caption{}
    \label{fig:Ours}
  \end{subfigure}

  \caption{\textbf{SGG Models.} (a) Detector-based model. (b) Query-based model. (c) Query-based model that additionally incorporates entity information derived from an object detector. (d) The proposed Dual-SGG method, which integrates detector-based and query-based reasoning mechanisms within a unified triplet decoder.}
  \label{fig:Introduction}
\end{figure}

To bridge this gap, we design a controlled experimental setup on the Visual Genome (VG) \cite{krishna2017visual} dataset to compare prediction behaviors from the perspective of detector-conditioned reachability.  To this end, we pre-train an external object detector that serves as a discriminator, determining whether a triplet is reachable for a detector-based SGG model under the original detector condition. Specifically, the ground-truth triplet set is split into two subsets. (1) \textbf{Det-T}, comprising triplets for which both the subject and object instances are successfully detected by the external object detector; and (2) \textbf{UDet-T}, comprising triplets for which at least one of the subject or object instances is not detected. Subsequently, the backbone, image encoder, and object decoder (when required) of the external object detector are employed and fine-tuned within three distinct SGG models. The first is a detector-based model (\cref{fig: Detector-based SGG}), constructed following \cite{im2024egtr}. The second is a query-based model (\cref{fig: Queries-based SGG}), implemented in accordance with \cite{khandelwal2022iterative}. The third is an additional query-based model (\cref{fig: Queries-based SGG with prior knowledge}) that leverages prior entity information extracted from the object detector, following \cite{cong2023reltr}. To control for confounding factors other than the underlying reasoning mechanisms, all three models are trained under an identical experimental protocol.
We evaluate the performance of each model on Det-T and UDet-T sets with micro-Recall, denoted as micro-DR (on Det-T) and micro-UDR (on UDet-T), in addition to standard aggregate metrics. The results are reported in \cref{tab:Introduction}. The detector-based model \cref{tab:Introduction} (a) achieves strong performance on Det-T but nearly collapses on UDet-T, demonstrating that fine-tuning the object detector in the SGG task does not fundamentally boost the detector-based model to capture triplets involving unreachable instances for the employed object detector. In contrast, query-based models \cref{tab:Introduction} (b, c) substantially improve micro-UDR. This indicates that query-based models can explore more triplets that, due to the detector constraint, are challenging for the detector-based model. However, the improvement in micro-UDR does not lead to better SGG performance. The absence of the explicit detector condition results in significant drops in micro-DR even when entity information is incorporated.
\begin{table}[t]
  \caption{\textbf{Pre-ablation study on the SGG models presented in \cref{fig:Introduction}.} The performance is evaluated on the VG validation dataset. All metrics are computed using the $\operatorname{Top}100$ predictions from each model. Here, micro-R, R, and mR correspond to micro-Recall, Recall, and mean-Recall, respectively, computed over the complete ground-truth set. The metrics micro-DR and micro-UDR represent micro-Recall computed on the corresponding specialized subsets.
  }
  \label{tab:Introduction}
  \centering
  \begin{tabular}{@{}cccccc@{}}
    \toprule
    Model & micro-R & micro-DR &micro-UDR &R&mR\\
    \midrule
    a  & 32.5 & 53.6 & 0.3& 33.4 & 10.9\\
    b & 30.7 & 47.3 &5.4 & 31.4& 9.6\\
    c & 31.8 & 48.8 & 5.8 & 32.4&10.3 \\

    d & 34.2 & 53.6 & 4.8 & 35.0 &11.2\\
  \bottomrule
  \end{tabular}
\end{table}

These empirical results suggest that detector- and query-based methods exhibit complementary predictive behaviors. Motivated by this complementarity, we seek a unified method that integrates both predictive behaviors. A straightforward solution would be to merge the detected triplets from detector- and query-based models by adopting the probabilistic ensembling \cite{chen2022multimodal} in the object detection task. However, compared to the object detection task, the triplet scores are collectively determined by three sub-probability distributions, which are non-trivial to calibrate. We therefore propose a \textbf{Dual-SGG} method, which models the mechanisms of detector- and query-based methods within a single triplet decoder, as illustrated in \cref{fig:Ours}. These two mechanisms are achieved by a dual-query design. The first group of queries is \textit{top-down} triplet queries (TD-Qs). Specifically, they detect triplets by conditioning on given entity pairs proposed by an entity pair selector (EPS). The second group of queries is called \textit{bottom-up} triplet queries (BU-Qs). All BU-Qs are established to globally explore triplets by originating from the image center and subsequently expanding outwards. With this approach, the final predictions encompass triplets based on both predictive behaviors. Since the predictions are obtained from a unique decoder, explicit post-hoc calibration among the predictions is not required. As demonstrated by \cref{tab:Introduction} (d), our proposed method maintains equal micro-DR to the detector-based model while increasing micro-UDR, thus leading to overall superior performance.

In summary, our contributions are as follows: We systematically and quantitatively compare the predictive behaviors of query- and detector-based SGG methods from the perspective of detector-conditioned reachability. Motivated by the results, we propose a method that integrates both detector-based and query-based reasoning mechanisms within a single triplet decoder to fuse their predictive behaviors. We conduct extensive experiments on three public datasets to demonstrate the effectiveness and efficiency of the proposed method.
\section{Related Work}
\subsubsection{Scene Graph Generation.} 
Early SGG works largely adopt the detector-based method that employs an off-the-shelf Faster R-CNN \cite{ren2016faster} to propose entities. Given a detected entity pair, \cite{lu2016visual} leverages entity categories, locations, visual representations, and language prior knowledge to predict the predicates between them. \cite{zellers2018neural} proposes utilizing global context information to support SGG. More recently,  \cite{zheng2023prototype, li2024leveraging} focus on improving predicate and triplet representations for predicate understanding. 

Query-based SGG methods \cite{teng2022structured, chen2024hydra} update a set of triplet queries to achieve SGG. Iterative-SGG \cite{khandelwal2022iterative} decomposes the triplet queries into subject, object, and predicate queries and models their dependencies with three transformer \cite{vaswani2017attention} decoders. \cite{cong2023reltr,li2022sgtr} propose introducing the entity information from an object detector into triplet query updating.
SpeaQ \cite{kim2024groupwise} adopts one-to-many and groupwise matching at the triplet level to enhance supervision during training. 
\subsubsection{Improve DETR Training with Auxiliary Queries.}
Due to one-to-one Hungarian matching \cite{kuhn1955hungarian}, only a few detections can be assigned positive labels during the training of DETR \cite{carion2020end}, which may impede training speed. Numerous efforts \cite{jia2023detrs, chen2023group, zhao2024ms, zhang2025mr} have been made to accelerate training by performing one-to-many matching strategies with auxiliary queries. However,  labels assigned through Hungarian matching may still pose optimization challenges. To address this issue, DN-DETR \cite{li2022dn} and DINO-DETR \cite{zhang2022dino} adopt dual-query designs. They introduce additional queries formed by perturbing ground-truth labels and train them with a denoising objective for stable supervision. For this purpose, these auxiliary queries are used only during training.

Inspired by the query grouping, we adopt a dual-query for SGG. To the best of our knowledge, this study is the first to propose fusing distinct predictive behaviors for SGG. The dual-query is designed for a fundamentally different purpose than the object detection work: not to stabilize training, but to explicitly model complementary prediction within a single end-to-end framework, replacing the need for multiple specialized models and complex fusion strategies.
\section{Method}
\label{sec:method}
\subsubsection{Preliminary.} SGG aims to predict a set of triplets $R=\{r_{k}\}=\{(s_{k},o_{k},p_{k})\}_{k=1}^{K}$ where $s_k,o_k, p_{k}$ denote the subject, object, and predicate, respectively. During inference, SGG computes a score for every triplet candidate and returns the $\operatorname{Top}K$ highest scoring triplets as the final predictions, \ie, $R=\operatorname{Top}K_{r}(S(r))$.
According to the reasoning mechanism, the scoring function \(S(r)\) of the detector-based method can be expressed as $S(r)=S_{d}(r\mid e_{s},e_{o},I)$, where $I$ represents the image features,  \(e_{s}\) and \(e_{o}\) denote the subject–object pair derived from the detected entity set \(E\), which is \(e_{s}, e_{o} \in E\). In contrast, for the query-based SGG method, the score function can be formulated as $S(r)=S_{q}(r \mid I)$.
As this study aims to preserve the predictive behaviors of both methods, we formulate the score function as follows:
\begin{align}
  S(r)=\max (S_{d}(r\mid e_{s},e_{o},I),\; S_{q}(r\mid I))
\label{infernce formulation}
\end{align}

However, if $d$ and $q$ in \cref{infernce formulation} are independent, directly applying a $\max$ operation to their scores may cause one branch to dominate due to score scale inconsistencies, which would undermine the intended complementarity between the two prediction sets. To avoid this issue, we integrate both reasoning mechanisms into a unified triplet decoder $\theta$ via a dual-query design, enabling joint optimization within a single framework. Let $\hat{R}=\{\hat{r}_x\}=\{(\hat{s}_x, \hat{o}_x, \hat{p}_x)\}$ denote the set of ground-truth triplets, while the objective of our method is:
\begin{align} 
\max_{\theta} \sum_{x=1}^{X} \left( \log S_{\theta}(\hat{r}_x \mid I) + \sum_{(s,o)}( m_x \log S_{\theta}(\hat{r}_x \mid e_s, e_o, I) )\right) \label{objective} 
\end{align}
This objective maximizes the log-likelihood of each ground-truth triplet under two conditions: (1) global image-conditioned prediction, corresponding to query-based reasoning, and (2) detector-conditioned prediction. The indicator variable $m_{x}$ activates the second term only when such a match exists, defined as:
\begin{align}
  m_{x}=\mathbf{1} [\operatorname{match}(\hat{s}_{x}, e_{s})\land \operatorname{match}(\hat{o}_{x}, e_{o})]
\label{indicator}
\end{align}
thereby ensuring that the detector-conditioned supervision is applied exclusively when the subject and object instances are reachable for the object detector.
\subsubsection{Model Overview.} 
\begin{figure}[tb]
  \centering
  \includegraphics[height=5.5cm]{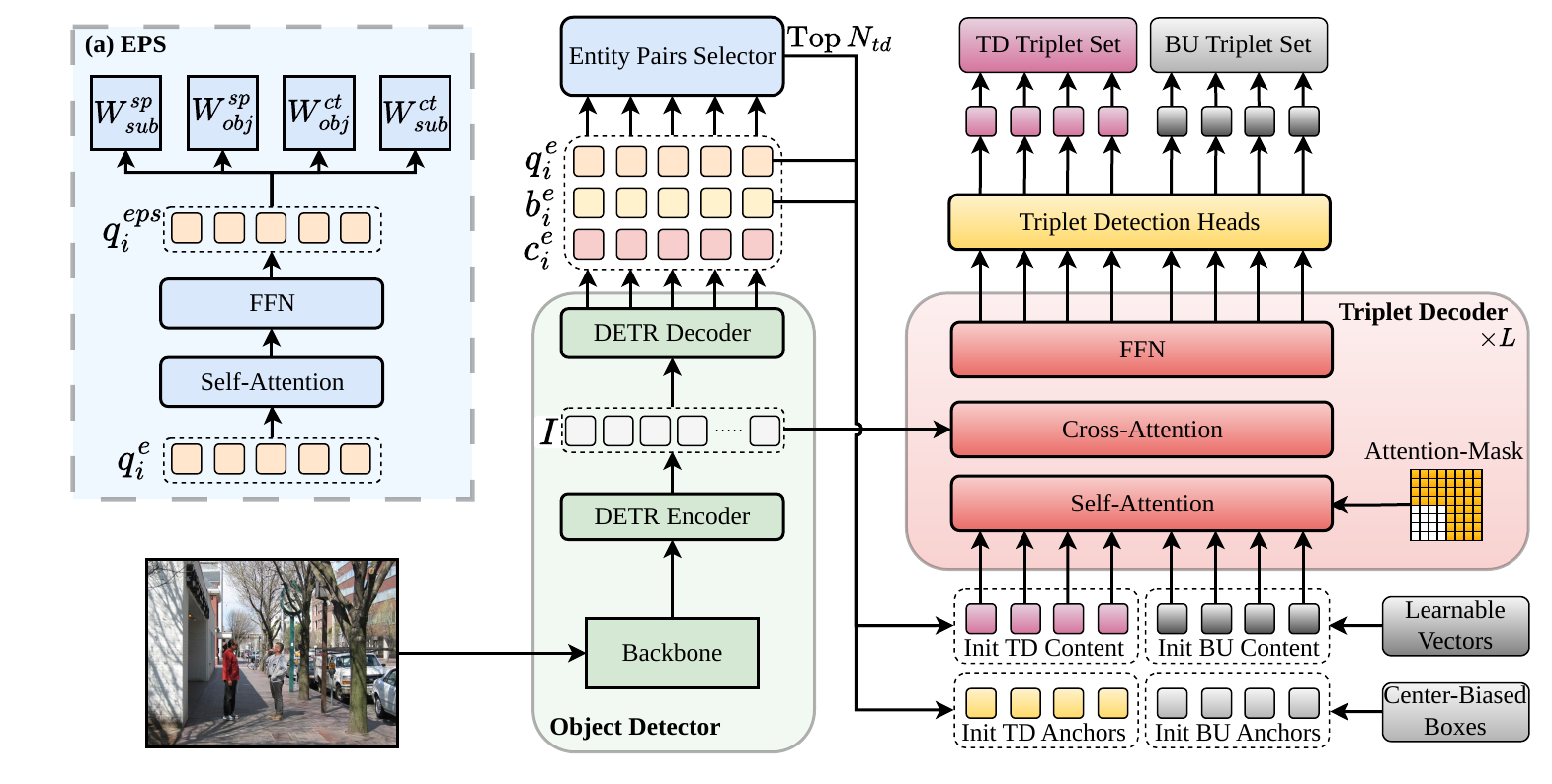}
  \caption{\textbf{Illustration of our Dual-SGG method.} It includes an object detector, an entity pairs selector (EPS), and a triplet decoder. FFN denotes the feed-forward network. TD content and TD anchors form TD-Qs, while BU content and BU anchors form BU-Qs. (a) presents architectural details of EPS.
  }
  \label{fig:dual-sgg}
\end{figure}

Our Dual-SGG is illustrated in \cref{fig:dual-sgg}. An object detector first extracts visual features and detects entities (see \cref{sec:object detector}). An Entity Pair Selector (EPS) selects subject-object pairs from these entities to initialize TD-Qs for detector-conditioned reasoning (see \cref{sec:eps}). Alongside TD-Qs, BU-Qs are initialized without any entity information for global image-conditioned prediction (see \cref{sec: initlaizrtion}). Both query types, including content embeddings and triplet anchors, are jointly updated via a unified triplet decoder (see \cref{sec: triplet_decoder}). Ultimately, the triplet detection heads are applied for triplet prediction.
\subsection{Object Detector}
\label{sec:object detector}
We adopt Deformable-DETR \cite{zhu2020deformable} as the object detector. Given an image, the encoder first extracts visual features $I$. Subsequently, the decoder generates a set of entity candidates, denoted as $E=\{e_{i}\}=\{(q^{e}_{i}, b^{e}_{i},c^{e}_{i})\}_{i=1}^{N_{e}}$, where $q^{e}_{i} \in \mathbb{R}^{d}$, $b^{e}_{i} \in \mathbb{R}^{4}$, and $c^{e}_{i}\in \mathbb{R}^{N_{c}}$ refer to the entity features, boxes, and category distributions, respectively.

\textbf{Object Detector Loss.} We follow \cite{zhu2020deformable} to train the object detector. The Hungarian matching is applied to search for an optimal assignment between $E$ and the ground-truth entity set by minimizing a global matching cost.
Based on this assignment, the ground-truth entity set is permuted and padded with \textit{no-object} labels for the unmatched detections. We define the permuted ground-truth set as $\hat{E}=\{\hat{e}_{i}\}=\{(\hat{b}^{e}_{i},\hat{c}^{e}_{i})\}_{i=1}^{N_{e}}$, where $\hat{b}^{e}_{i} \in \mathbb{R}^{4}$ and $\hat{c}^{e}_{i}\in \mathbb{R}^{N_{c}}$ indicate the target boxes and one-hot class labels of entities.  The object detector loss includes a focal loss \cite{lin2017focal} for entity classification and L1 plus GIoU losses for box regression:
\begin{align}
\mathcal{L}^{e}=\sum_{i=1}^{N_{e}}[\lambda^{cls}\mathcal{L}^{cls}(c_{i}^{e}, \hat{c}^{e}_{i})+\lambda^{box}\mathcal{L}^{box}(b^{e}_{i}, \hat{b}^{e}_{i})+\lambda^{giou}\mathcal{L}^{giou}(b^{e}_{i}, \hat{b}^{e}_{i})]
\end{align}
Building upon the aligned entity set $\hat{E}$, we further construct a predicate matrix ${\hat{P}^{e}} \in \mathbb{R}^{N_{e} \times N_{e} \times N_{p}}$. For each detected entity pair $(i,j)$, $\hat{P}^e_{ij}$ is a one-hot vector indicating the ground truth predicate between entity $i$ and $j$. Note that the matrix ${\hat{P}^{e}}$ is used for training the triplet decoder and the EPS.
\subsection{Entity Pairs Selector (EPS)}
\label{sec:eps}
TD-Qs are initialized with the detected entity pairs to introduce the detector condition. However, enumerating all possible pairs results in $N_{e}^2$ queries, which leads to significant computational overhead. To reduce this cost, we employ an Entity Pair Selector (EPS) that evaluates both the content and spatial compatibility of entity pairs and selects the $\operatorname{Top}N_{td}$ candidates for TD-Qs initialization. The content score favors entity pairs that correspond to ground-truth triplets, providing stable positive supervision for the subsequent training of triplet prediction. The spatial score emphasizes pairs that are spatially similar to content-positive candidates, providing informative hard negative samples.

Unlike prior work \cite{jung2023devil, wang2024multi, qu2026salience}, which uses intricate modules to evaluate detected triplets and adjust their scores, EPS serves here as a selector rather than a post-hoc scoring mechanism. While not directly determine final predictions, EPS enables a computationally efficient design, as shown in \cref{fig:dual-sgg} (a). EPS includes a self-attention layer and a FFN for updating the entity features $\{q^{e}_{i}\}_{i=1}^{N_{e}}$, yielding $\{q^{eps}_{i}\}_{i=1}^{N_{e}}$. The spatial logits $S^{sp} \in \mathbb{R}^{N_e \times N_{e}}$ and content logits $S^{ct} \in \mathbb{R}^{N_e \times N_{e}}$ are obtained by performing a dot-product as follows:
\begin{align}
S^{sp}= \{q^{eps}_{i}\}W^{sp}_{sub} \cdot 
 (\{q^{eps}_{i}\}W_{obj}^{sp})^{T} \qquad S^{ct}&= \{q^{eps}_{i}\}W^{ct}_{sub} \cdot 
 (\{q^{eps}_{i}\}W_{obj}^{ct})^{T} 
\end{align}
where $W^{sp}_{sub}, W_{obj}^{sp}, W^{ct}_{sub}, W_{obj}^{ct}$ are linear projectors. Ultimately, the final scores for entity pairs  $S^{eps} \in \mathbb{R}^{N_e \times N_{e}}$ are calculated as follows:
\begin{equation}
S^{eps} =
\begin{cases}
\operatorname{sigmoid}(S^{sp})\times \operatorname{sigmoid}(S^{ct}), & \text{Training} \\
\operatorname{sigmoid}(S^{ct}), & \text{Inference}
\end{cases}
\end{equation}

\textbf{EPS Loss.} The EPS loss consists of two focal losses: spatial loss $\mathcal{L}^{sp}$ and content loss $\mathcal{L}^{ct}$.
\begin{align}
\mathcal{L}^{eps}=\mathcal{L}^{sp}(S^{sp}, \hat{S}^{sp})+ \mathcal{L}^{ct}(S^{ct},\hat{S}^{ct})
\label{equ: EPS}
\end{align}
The content labels $\hat{S}^{ct} \in \mathbb{R}^{N_{e} \times N_{e}} $ are derived by converting the predicate matrix $\hat{P}^{e}$ introduced in \cref{sec:object detector} into a binary matrix:
\begin{equation}
\hat{S}^{ct}_{ij} =
\begin{cases}
1, & \text{If}\; \sum_{k=1}^{N_{p}} \hat{P}^{e}_{ijk} > 0 \\
0, & \text{Otherwise}
\end{cases}
\end{equation}
Rather than using the locations of ground-truth triplets, we use the locations of detected entity pairs with content-positive labels to assign spatial labels, as shown in \cref{eq:spatial_label}. This ensures that all spatial-positive pairs exhibit spatial configurations similar to the content-positive samples:
\begin{equation}
\hat{S}^{sp}_{ij} =
\begin{cases}
1, & \text{if } 
\mathrm{IoU}(b^{e}_i, b^{e}_x) > 0.5
\land
\mathrm{IoU}(b^{e}_j, b^{e}_y) > 0.5
\land
\hat{S}^{ct}_{xy} = 1, \\
0, & \text{otherwise.}
\end{cases}
\label{eq:spatial_label}
\end{equation}
where $b^{e}_{i}, b^{e}_{j}, b^{e}_{x},b^{e}_{y} \in E$ are the entity boxes produced by the object detector.
\subsection{Triplet Queries Initialization}
\label{sec: initlaizrtion}
Inspired by DAB-DETR \cite{liu2022dab}, our TD-Qs and BU-Qs consist of content embeddings and triplet anchors. The triplet anchors are introduced to guide the attention of the triplet decoder in detecting triplets in two different patterns.

\textbf{TD-Qs Initialization.} To impose the detector condition, TD-Qs are constructed with the selected pairs of entities. Specifically, the TD-Qs content embeddings and triplet anchors are initialized by concatenating the corresponding entity features and bounding boxes, respectively. Given $S^{eps} \in \mathbb{R}^{N_{e} \times N_{e}}$, $\operatorname{Top}N_{td}$ entity pairs are selected. Let $\tilde{s}_{(i)}$ and $\tilde{o}_{(i)}$  denote the indices of the entities involved in the selected pairs by EPS. The content embedding $\{q^{td}_{i}\}_{i=1}^{N_{td}}$ and triplet anchors $\{a^{td}_{i}\}_{i=1}^{N_{td}}$ are initialized as follows:
\begin{align}
\{q^{td}_{i}\}=\operatorname{Concat}(\{q^{e}_{\tilde{s}_{(i)}}\},\{q^{e}_{\tilde{o}_{(i)}}\}), \qquad \{a^{td}_{i}\}=\operatorname{Concat}(\{b^{e}_{\tilde{s}_{(i)}}\},\{b^{e}_{\tilde{o}_{(i)}}\})
\end{align}
where $q^{td}_{i} \in \mathbb{R}^{2d}$ and $a^{td}_{i} \in \mathbb{R}^{8}$.

\textbf{BU-Qs Initialization.} For BU-Qs, the content embeddings $\{q^{bu}_{i}\}_{i=1}^{N_{bu}}$, where $q^{bu}_{i} \in \mathbb{R}^{2d}$, are randomly initialized with a set of learnable vectors. 
The anchors of BU-Qs, denoted as $\{a^{bu}_{i}\}_{i=1}^{N_{bu}}$, where $a^{bu}_{i} \in \mathbb{R}^{8}$, are initialized with a set of pairwise frozen boxes centered in the image. These center-biased anchors encourage the decoder to initially focus on the central region and gradually expand its attention outwards across decoder layers. Consequently, the BU-Qs capture triplets that are not reachable for the TD-Qs due to the detector constraint. 
\subsection{Triplet Decoder}
\label{sec: triplet_decoder}
The triplet decoder consists of $L$ transformer \cite{vaswani2017attention} blocks. To reduce computation, we adopt a single-branch decoder that reasons over complete triplets. This design can be extended to a multi-branch version, where each branch models a triplet component, as in \cite{khandelwal2022iterative, cong2023reltr, fu2025hybrid}. The decoder takes the initialized TD-Qs and BU-Qs as input and updates them jointly through stacked transformer blocks. Triplet detection heads are then applied to the updated queries, producing two triplet sets denoted as $R^{td}=\{r^{td}_{i}\}=\{(s^{td}_{i},o^{td}_{i},p^{td}_i)\}_{i=1}^{N_{td}}$ and $R^{bu}=\{r^{bu}_{i}\}=\{(s^{bu}_{i},o^{bu}_{i},p^{bu}_i)\}_{i=1}^{N_{bu}}$, respectively.  To prevent the rich prior in TD-Qs from leaking into BU-Qs and causing a significant dependency of $R^{bu}$ on $R^{td}$, we use self-attention masks \cite{vaswani2017attention} to restrict the information flow from TD-Qs to BU-Qs in each transformer's self-attention layer.

\textbf{TD-Loss.} To ensure $R^{td}$ predicted by TD-Qs follows the provided detector condition, we construct the labels for triplets detected with TD-Qs, denoted as $\hat{R}^{td}=\{\hat{r}^{td}_{i}\}=\{(\hat{s}^{td}_{i},\hat{o}^{td}_{i},\hat{p}^{td}_i)\}_{i=1}^{N_{td}}$, based on the labels of their initial entities obtained from object detector training. Specifically, given the ground-truth labels  $\hat{E}=\{\hat{e}_{i}\}=\{(\hat{b}^{e}_{i},\hat{c}^{e}_{i})\}_{i=1}^{N_{e}}$ of the detected entities introduced in \cref{sec:object detector}, the labels of subjects and objects are extracted as  $\{\hat{s}^{td}_{i}\}=\{(\hat{b}^{e}_{\tilde{s}_{(i)}}, \hat{c}^{e}_{\tilde{s}_{(i)}})\}_{i=1}^{N_{td}}$ and $\{\hat{o}^{td}_{i}\}=\{(\hat{b}^{e}_{\tilde{o}_{(i)}}, \hat{c}^{e}_{\tilde{o}_{(i)}})\}_{i=1}^{N_{td}}$, respectively, where $\tilde{s}_{(i)}$ and $\tilde{o}_{(i)}$ are the indices of entities involved by the selected entity pairs, as introduced in \cref{sec: initlaizrtion}. The predicate labels are extracted from $\hat{P}^{e}$ defined as $\{\hat{p}^{td}_{i}\}=\{\hat{P}^{e}_{\tilde{s}_{(i)}\tilde{o}_{(i)}}\}_{i=1}^{N_{td}}$. The entire objective of TD-Qs is as follows:
\begin{align}
\mathcal{L}^{TD}=\sum_{i}^{N_{td}}[\mathcal{L}^{TD}_{sub}(s^{td}_{i},\hat{s}^{td}_{i})+\mathcal{L}^{TD}_{obj}(o^{td}_{i},\hat{o}^{td}_{i})+\mathcal{L}^{TD}_{p}(p^{td}_{i},\hat{p}^{td}_{i})]
\label{loss_td}
\end{align}
where $L^{TD}_{p}$ is a focal loss, and $\mathcal{L}^{TD}_{sub}$ and $\mathcal{L}^{TD}_{obj}$ are identical and share the same loss weights with the object detector loss. For instance:
\begin{align}
\mathcal{L}^{TD}_{sub}(s^{td}_{i},\hat{s}^{td}_{i})=\lambda^{cls}\mathcal{L}^{cls}(s^{td}_{i,c}, \hat{s}^{td}_{i,c})+\lambda^{box}\mathcal{L}^{box}(s^{td}_{i,b}, \hat{s}^{td}_{i, b})+\lambda^{giou}\mathcal{L}^{giou}(s^{td}_{i,b}, \hat{s}^{td}_{i,b})
\label{loss_td_entity}
\end{align}
$s^{td}_{i,c}, \hat{s}^{td}_{i,c}$ refer to the categories and $s^{td}_{i,b}, \hat{s}^{td}_{i,b}$ indicate the boxes of corresponding subjects. Note that the TD-Qs may not be trained on all ground-truth triplets due to the selection process of EPS.

\textbf{BU-Loss.} BU-Qs are supervised following the standard query-based SGG method, where labels are assigned by applying Hungarian matching between detected triplets and ground-truth triplets in a one-to-one manner, as in \cite{khandelwal2022iterative}. However, in our approach, the predicate cost is excluded from the matching cost calculation, as its computational overhead outweighs its empirical benefit. Ultimately, given the labels for the detected triplets, the calculation of BU-loss, denoted as $\mathcal{L}^{BU}$, is analogous to \cref{loss_td} and \cref{loss_td_entity}.
\subsection{Training and Inference}
\textbf{Training.} In summary, the entire loss function of our method is:
\begin{align}
\mathcal{L}=\mathcal{L}^{e}+\mathcal{L}^{eps}+\mathcal{L}^{TD}+\mathcal{L}^{BU}
\label{eq:overall objective}
\end{align}
where  \(\mathcal{L}^{TD}\) and  \(\mathcal{L}^{BU}\) realize $ \sum_{(s,o)}( m_x \log S_{\theta}(\hat{r}_x \mid e_s, e_o, I) )$ and $\log S_{\theta}(\hat{r}_x \mid I)$ in \cref{objective}, respectively. \(\mathcal{L}^{e}\) and \(\mathcal{L}^{eps}\) further ensure high-quality entity-pair proposals for TD-Qs initialization. Whilst the present work introduces additional loss terms in comparison to previous studies, all newly added losses are assigned a unit weight to isolate the influences from hyperparameter tuning. \\
\textbf{Inference.} In the inference stage, the predictions $R^{td}$ and $R^{bu}$ are concatenated to form the final prediction set $R=\{r_{i}\}_{i=1}^{N_{td}+N_{bu}}$. In addition, Non-Maximum Suppression (NMS) is applied to serve as a practical implementation of the max operation in \cref{infernce formulation} for removing duplicate triplets, as in \cite{kim2024groupwise, khandelwal2022iterative, li2024leveraging, wang2024multi, cong2023reltr}. Ultimately, $\operatorname{Top}K$ triplets are selected based on the triplet scores, which are calculated by multiplying the subject, object, and predicate scores together.

\section{Experiments}
\subsection{Experimental Setup}
This section contains the descriptions of datasets, evaluation metrics, and implementation details.

\textbf{Datasets.} The experiments are conducted on the following datasets: \\
\textit{Visual Genome (VG)} \cite{krishna2017visual} includes 57,723 training, 5,000 validation, and 26,446 test images. It defines 150 entity and 50 predicate categories. \\
\textit{Open Images V6 (OIv6)} \cite{kuznetsova2020open} consists of 126,368 training, 1,813 validation, and 5,322 test images. It features 601 entity and 30 predicate categories. \\
\textit{GQA-200} \cite{hudson2019gqa} contains 27,623 training, 5,000 validation, and 8,208 test images, comprising 200 entity and 100 predicate categories.

\textbf{Evaluation Metrics.} We evaluate our method on the scene graph detection (SGDet) task, following \cite{teng2022structured, im2024egtr, zellers2018neural, li2024leveraging, wang2024multi}. On VG and GQA-200, Recall@K (R@K), mean Recall@K (mR@K), and their harmonic mean F@K are reported under the graph constraint proposed by \cite{zellers2018neural}. For OIv6, we report mR@50, micro-Recall@50 (micro-R@50), weighted mean AP of triplets (wmAPrel), weighted mean AP of phrases (wmAPphr), and a final score defined as $\text{score}=0.2\times\text{micro-R@50}+0.4\times\text{wmAPrel}+0.4\times\text{wmAPphr}$.

\textbf{Implementation Details.} For all three datasets, we adopt the same training setup and hyperparameters. The training consists of two parts. First, we pre-train the object detector (Deformable-DETR \cite{zhu2020deformable}) with a ResNet50 \cite{he2016identity} backbone and 200 entity queries ($N_e$). The loss weights $\lambda^{cls}$, $\lambda^{box}$, and $\lambda^{giou}$ are 2, 5, and 2. Next, the EPS and triplet decoder are jointly trained while fine-tuning the pre-trained object detector. We set $N_{td}=300$, $N_{bu}=800$, and $L=4$. We use AdamW \cite{loshchilov2017decoupled} as an optimizer with an initial learning rate of $10^{-4}$ and a weight decay of $10^{-4}$. The batch size is 16 in total. This second part is trained for 25 epochs, with the learning rate reduced by a factor of 0.1 at epoch 20. All experiments run on a single NVIDIA RTX A6000 GPU.
\subsection{Main Results}
\begin{table}[t]
\caption{\textbf{Graph-Constraint results on the VG test set.} Methods are grouped by backbone into \textit{X101-FPN} \cite{lin2017feature} (top) and \textit{ResNet} \cite{he2016identity} (bottom). In each group, the best and second-best results are highlighted in \textbf{bold} and \underline{underline}. $\checkmark$ denotes methods with debiasing strategies (\ie Unbiased-SGG). LA indicates logit adjustment \cite{menon2020long}}.
  \centering
  \resizebox{\linewidth}{!}{
  \begin{tabular}{@{} c l|c c|c c |c c |c c @{}}
    \toprule
    Unbiased &Method & \#params(M)& FPS & R@50 & R@100 & mR@50 & mR@100 & F@50 & F@100 \\
    \midrule
    \space&Motifs \cite{zellers2018neural}{\scriptsize \textcolor{gray}{[CVPR'18]}}&369.9&1.9
    & 32.1 & 36.9 & 5.5 & 6.8 & 9.4 & 11.5 \\
    \space&VCTree \cite{tang2019learning}{\scriptsize \textcolor{gray}{[CVPR'19]}}&361.5&0.8
    & 31.8 & 36.1 & 6.6 & 7.7 & 10.9 & 12.7 \\
    $\checkmark$&BGNN \cite{li2021bipartite}{\scriptsize \textcolor{gray}{[CVPR'21]}}&341.9&1.7
    & 31.0 & 35.8 & 10.7 & 12.6 & 15.9 & 18.6 \\
    $\checkmark$&DT2-ACBS \cite{desai2021learning}{\scriptsize \textcolor{gray}{[ICCV'21]}}&-&-
    & 15.0 & 16.3 & \textbf{22.0} &  \textbf{24.4} & 17.8 & 19.5 \\
    $\checkmark$&IETrans \cite{zhang2022fine}{\scriptsize \textcolor{gray}{[ECCV'22]}}&369.9&1.9
    & 23.5 & 27.3 & 15.7 & 18.2 & 18.8 & 21.8 \\
    $\checkmark$&SHA \cite{dong2022stacked}{\scriptsize \textcolor{gray}{[CVPR'22]}}&-&-
    & 14.9 & 18.2 & 17.9 & 20.9 & 16.3 & 19.5 \\
    \space&SSR-CNN \cite{teng2022structured}{\scriptsize \textcolor{gray}{[CVPR'22]}}&274.3&3.1
    & \underline{33.5} & \underline{38.4} & 8.6 & 10.3 & 13.7 & 16.2 \\
    $\checkmark$&SSR-CNN+LA$^{*}$ \cite{teng2022structured}{\scriptsize \textcolor{gray}{[CVPR'22]}}&274.3&3.1
    & 23.7 & 27.3 & 18.6 & 22.5 & \textbf{20.8} & \textbf{24.7} \\
    \space&PE-Net \cite{zheng2023prototype}{\scriptsize \textcolor{gray}{[CVPR'23]}}&-&-
    & 32.4 & 36.9 & 8.9 & 11.0 & 14.0 & 16.9 \\
    \space&DRM \text{w/o} DKT \cite{li2024leveraging}{\scriptsize \textcolor{gray}{[CVPR'24]}}&-&-
    & \textbf{34.0} & \textbf{38.9} & 9.0 & 11.2 & 14.2 & 17.4 \\
    $\checkmark$&DRM \cite{li2024leveraging}{\scriptsize \textcolor{gray}{[CVPR'24]}}&-&-
    & 19.0 & 22.9 & \underline{20.4} & \underline{24.1} & \underline{20.8} & \underline{23.5} \\
    $\checkmark$&RA-SGG \cite{yoon2025ra}{\scriptsize \textcolor{gray}{[AAAI'25]}}&-&-
    & 26.0 & 30.3 & 14.4 & 17.1 & 18.5 & 21.9 \\
    \midrule
    \midrule
    $\checkmark$&SGTR \cite{li2022sgtr}{\scriptsize \textcolor{gray}{[CVPR'22]}}&117.1&6.2
    & 25.1 & 26.6 & 12.0 & 14.6 & 16.2 & 18.9 \\
    \space&RelTR \cite{cong2023reltr}{\scriptsize \textcolor{gray}{[TPAMI'22]}}&63.7&13.4
    & 27.5 & 30.7 & 10.8 & 12.3 & 15.5 & 17.6 \\
    \space&Relationformer \cite{shit2022relationformer}{\scriptsize \textcolor{gray}{[ECCV'22]}}&92.9&8.5
    & 28.4& 31.3 &9.3 &10.7 & 14.0 & 15.9 \\
    \space&ISG \cite{khandelwal2022iterative}{\scriptsize \textcolor{gray}{[NIPS'22]}}&93.5&6.0
    & 29.7 & 32.1 & 8.0 & 8.8 & 12.6 & 13.8 \\
    $\checkmark$&Mg-RMPN \cite{wang2024multi}{\scriptsize \textcolor{gray}{[ECCV'24]}}&-&-
    & 29.1 & 33.5 & 14.4 & 17.3 & 19.3 & 22.8 \\
    \space&EGTR \cite{im2024egtr}{\scriptsize \textcolor{gray}{[CVPR'24]}}&42.5&14.7
    & 30.2 & 34.3 &7.9 & 10.1 & 12.5 & 15.6 \\
    $\checkmark$&EGTR+LA \cite{im2024egtr}{\scriptsize \textcolor{gray}{[CVPR'24]}}&42.5&14.7
    & 24.2 & 26.7 &17.1 & 21.4 & 20.0 & 23.8 \\
    $\checkmark$&Hydra-SGG \cite{chen2024hydra}{\scriptsize \textcolor{gray}{[ICLR'25]}}&67.6&5.3
    & 28.4 & 33.1 &16.0 & 19.7 & 20.5 & 24.7 \\
    $\checkmark$&Salience-SGG \cite{qu2026salience}{\scriptsize \textcolor{gray}{[WACV'26]}}&77.7&13.3
    & 28.8& 33.4 &18.0 &21.6 & 22.1 & 26.2 \\
    \space&Dual-SGG (Ours)&84.7&14.0
    & \textbf{33.5}& \textbf{38.5} &10.3 &12.3 & 15.8 & 18.6 \\
    $\checkmark$&Dual-SGG+LA \scriptsize  {$\tau=0.2$} (Ours)&84.7&14.0
    & \underline{30.2}& \underline{35.2} &\underline{18.7} &\underline{22.2} & \underline{23.1} & \underline{27.3} \\
    $\checkmark$&Dual-SGG+LA \scriptsize  {$\tau=0.3$} (Ours)&84.7&14.0
    & 25.3& 30.0 &\textbf{22.2} &\textbf{25.2} & \textbf{23.6} & \textbf{27.4} \\
    \bottomrule
  \end{tabular}
  }
  \label{tab:vg}
\end{table}
\begin{table}[t]
\caption{\textbf{Graph-Constraint results on the GQA-200 test set.}}
  \centering
  \resizebox{0.7\linewidth}{2cm}{
  \begin{tabular}{@{} c l|c c |c c |c c @{}}
    \toprule
    Unbiased &Method  & R@50 & R@100 & mR@50 & mR@100 & F@50 & F@100 \\
    \midrule
    \space&VTransE \cite{zhang2017visual}{\scriptsize \textcolor{gray}{[CVPR'17]}} & 27.2 & 30.7 & 5.8 & 6.6 & 9.6 & 10.9 \\
    \space&Motifs \cite{zellers2018neural}{\scriptsize \textcolor{gray}{[CVPR'18]}}
    & \textbf{28.9} & \textbf{33.1} & 6.4 & 7.7 & 10.5 & 12.5 \\
    \space&VCTree \cite{tang2019learning}{\scriptsize \textcolor{gray}{[CVPR'19]}}
    & \underline{28.3} & \underline{31.9} & 6.5 & 7.4 & 10.6 & 12.0 \\
    $\checkmark$& SHA \cite{dong2022stacked}{\scriptsize \textcolor{gray}{[CVPR'22]}}
    & 14.8 & 17.9 & \underline{17.8} & \underline{20.1} & \underline{16.2} & \underline{18.9} \\
    $\checkmark$&DRM \cite{li2024leveraging}{\scriptsize \textcolor{gray}{[CVPR'24]}}
    & 18.6 & 21.7 & \textbf{18.9} & \textbf{21.0} & \textbf{18.7} & \textbf{21.3} \\
    \midrule
    $\checkmark$&Mg-RMPN \cite{wang2024multi}{\scriptsize \textcolor{gray}{[ECCV'24]}}
    & 23.2 & 25.7 & 12.8 & 14.5 & 16.5 & 18.5 \\
    $\checkmark$&Hydra-SGG \cite{chen2024hydra}{\scriptsize \textcolor{gray}{[ICLR'25]}}
    & 22.8 & 26.5 &12.7 & 15.9 & 16.3 & 19.9 \\
    $\checkmark$&Salience-SGG \cite{qu2026salience}{\scriptsize \textcolor{gray}{[WACV'26]}}
    & 23.6& 26.6 &16.2 &18.4 & 19.2 & 21.7 \\
    \space&Dual-SGG (Ours)
    & \textbf{29.4}& \textbf{33.3} &9.1 &10.4 & 13.9 & 15.8 \\
    $\checkmark$&Dual-SGG+LA \scriptsize  {$\tau=0.2$} (Ours)
    & \underline{26.2}& \underline{29.9} &\underline{17.1} &\underline{19.3} & \textbf{20.7} & \textbf{23.4} \\
    $\checkmark$&Dual-SGG+LA \scriptsize  {$\tau=0.3$} (Ours)
    & 22.0& 25.7 &\textbf{19.0} &\textbf{21.1} & \underline{20.4} & \underline{23.2} \\
    \bottomrule
  \end{tabular}
  }
  \label{tab:gqa}
\end{table}
\begin{table}[t]
\caption{\textbf{Results on the OIv6 test set.}}
  \centering
  \resizebox{0.7\linewidth}{2cm}{
  \begin{tabular}{@{} l|c c |c c |c@{}}
    \toprule
    Method  & mR@50 & micro-R@50 & wmAPrel & wmAPphr & score \\
    \midrule
    BGNN \cite{li2021bipartite}{\scriptsize \textcolor{gray}{[CVPR'21]}}&40.5&75.0&33.5&34.2&42.1\\
    RU-Net \cite{lin2022ru}{\scriptsize \textcolor{gray}{[CVPR'22]}}&-&\textbf{76.9}&35.4&34.9&43.5\\
    SSR-CNN \cite{teng2022structured}{\scriptsize \textcolor{gray}{[CVPR'22]}}&42.8&\underline{76.7}&\textbf{41.5}&\textbf{43.6}&\textbf{49.4}\\
    PE-Net \cite{zheng2023prototype}{\scriptsize \textcolor{gray}{[CVPR'23]}}&-&76.5&36.6&37.4&44.9\\
    SQUAT \cite{jung2023devil}{\scriptsize \textcolor{gray}{[ICCV'23]}}&-&75.8&34.9&35.9&43.5\\
    DRM \cite{li2024leveraging}{\scriptsize \textcolor{gray}{[CVPR'24]}}&-&75.9&\underline{40.5}&\underline{41.4}&\underline{47.9}\\
    \midrule
    SGTR \cite{li2022sgtr}{\scriptsize \textcolor{gray}{[CVPR'22]}}&42.6&59.9&37.0&38.7&42.3\\
    RelTR \cite{cong2023reltr}{\scriptsize \textcolor{gray}{[TPAMI'22]}}&-&71.7&37.2&37.5&43.0\\
    Mg-RMPN \cite{wang2024multi}{\scriptsize \textcolor{gray}{[ECCV'24]}}&45.5&77.8&35.5&36.4&43.6\\
    Hydra-SGG \cite{chen2024hydra}{\scriptsize \textcolor{gray}{[ICLR'25]}}&-&76.1&42.8&44.3&50.1\\
    Salience-SGG \cite{qu2026salience}{\scriptsize \textcolor{gray}{[WACV'26]}}&48.0& \underline{78.1}&\underline{45.6}&\underline{44.9}&\underline{51.8}\\
    Dual-SGG (Ours)&48.0& \textbf{79.6}&\textbf{46.0}&\textbf{46.5}&\textbf{52.9}\\
    \bottomrule
  \end{tabular}
  }
  \label{tab:OIv6}
\end{table}
We evaluate our method by comparing it with previous SGG models, including Motifs \cite{zellers2018neural}, SSR-CNN \cite{teng2022structured}, PE-Net \cite{zheng2023prototype}, DRM w/o DKT \cite{li2024leveraging}, RelTR \cite{cong2023reltr}, EGTR \cite{im2024egtr}, and ISG \cite{khandelwal2022iterative}. To evaluate Dual-SGG on the long-tailed distribution problem, we apply logit adjustment (LA) \cite{menon2020long} as a debiasing strategy. The result is compared with other Unbiased-SGG models, such as BGNN \cite{li2021bipartite}, SHA \cite{dong2022stacked}, IETrans \cite{zhang2022fine}, SSR-CNN+LA \cite{teng2022structured}, DRM \cite{li2024leveraging}, EGTR+LA \cite{im2024egtr}, and Salience-SGG \cite{qu2026salience}. \textbf{Note}: We further compare with VETO \cite{sudhakaran2023vision}, SpeaQ \cite{kim2024groupwise}, and HQSG \cite{fu2025hybrid} without graph constraints in the \textbf{suppl. material}.

\textbf{VG Dataset.} \cref{tab:vg} presents the comparison results on the VG test set. Among SGG models with a ResNet \cite{he2016identity} backbone, Dual-SGG achieves the best results, outperforming the second-best EGTR in all metrics. Despite being heavier, our model achieves a similar inference speed to EGTR (14.0 FPS vs 14.7 FPS), as EGTR employs an MLP performing on $N_e^2$ detected entity pairs. RelTR exhibits close performance in mR@K and F@K to our method but has a significant lag in R@K of 6.0/7.8. Our method achieves comparable performance to DRM w/o DKT, which employs X101-FPN as the backbone. Specifically, the relative differences between Dual-SGG and DRM w/o DKT are -0.5/-0.4 in R@K, 1.3/1.1 in mR@K, and 1.6/1.2 in F@K. However, our Dual-SGG is much lighter and faster than DRM w/o DKT, as the X101-FPN backbone alone requires more parameters than our entire model. Regarding Unbiased-SGG, the debiasing strategy in Dual-SGG+LA leads to a relatively smaller degradation in R@K compared to SSR-CNN+LA and EGTR+LA. This is attributed to the improved exploration of subject-object instances by our Dual-SGG, which is less sensitive to predicate understanding. Consequently, our Dual-SGG+LA at $\tau=0.3$ achieves state-of-the-art performance, attaining F@K of 23.6/27.4 and mR@K of 22.2/25.2 simultaneously.

\textbf{GQA-200 Dataset.} The results on the GQA-200 test dataset are shown in \cref{tab:gqa}. Consistent with the observations for VG, Dual-SGG attains the best performance among SGG models, including VTransE, Motifs, and VCTree. Furthermore, compared with other Unbiased-SGG models, our Dual-SGG+LA at $\tau = 0.3$ establishes new state-of-the-art results, achieving mR@K of 19.0/21.1 and F@K of 20.4/23.2 simultaneously.

\textbf{OIv6 Dataset.} \cref{tab:OIv6} demonstrates the effectiveness of our method on the OIv6 dataset. Dual-SGG outperforms previous methods under all metrics.
\subsection{Ablation Studies}
In the ablation studies, all results are produced on the VG test dataset. The analysis of hyperparameters and efficiency is provided in the \textbf{suppl. material}.
\begin{table}[t]  
 \centering
\caption{\textbf{Ablation study on BU-Qs and TD-Qs.} Baseline refers to the model shown in \cref{fig: Detector-based SGG}. $*$ indicates that the model employ Swin-B as backbone and $N_{e}$ is set as 900.}
 
  \resizebox{\linewidth}{!}{
  \begin{tabular}{
    @{} c |c c|c c c 
    @{}
  }
    \toprule
    Models&R@50/100&mR@50/100& micro-R@50/100& micro-DR@50/100&micro-UDR@50/100 \\
    \midrule
    Baseline&31.6/36.0&9.1/11.2&30.2/34.8&46.7/53.8&0.2/0.3 \\
    Dual-SGG w/o BU-Qs&31.9/36.3&8.6/10.3&30.4/35.2&46.3/53.4&1.5/2.2 \\
    Dual-SGG w/o TD-Qs&31.3/34.7&10.0/11.6&30.0/33.7&44.1/49.0&4.5/5.9 \\
    Dual-SGG&33.5/38.5&10.3/12.3&31.8/36.9&47.3/54.3&3.7/5.2 \\
    \midrule
    \midrule
    Dual-SGG$^{*}$ w/o BU-Qs&37.1/42.7&10.1/13.0&35.2/40.6&46.5/53.6&1.5/1.7 \\
    Dual-SGG$^{*}$ w/o TD-Qs&35.3/38.5&11.3/13.0&33.6/37.2&43.4/47.6&4.4/6.1 \\
    Dual-SGG$^{*}$&38.8/44.6&11.7/14.8&36.3/42.5&47.2/54.9&3.8/5.7 \\
    \bottomrule
  \end{tabular}
  }
  \label{tab:instance-level detectability conditions SGG}
\end{table}
\begin{table}[t]  
 \caption{\textbf{Ablation study on complementarity}. The top part indicates the performances of specialized models. The bottom part shows the results from various fusion strategies.}
  \centering
  \resizebox{\linewidth}{!}{
  \begin{tabular}{
    @{} c |c c |c c c
    @{}
  }
    \toprule
Models&R@50/100&mR@50/100& micro-R@50/100& micro-DR@50/100&micro-UDR@50/100 \\
\midrule
    Dual-SGG w/o BU-Qs&31.7/36.1&8.4/9.9&30.4/34.9&46.3/53.1&1.5/1.8 \\
    Dual-SGG w/o TD-Qs&30.8/33.8&9.4/11.0&29.5/32.8&43.8/48.2&3.7/4.8 \\
\midrule
\midrule
    Max&31.7/36.3&9.0/11.2&28.9/34.1&43.1/50.3&3.2/4.7 \\
     Average&17.6/29.7&5.3/9.3&17.1/28.9&25.3/42.7&2.2/4.1 \\
     Normalization&32.2/37.3&9.6/11.8&30.7/36.0&45.8/53.2&3.3/4.9 \\
    Dual-SGG &33.5/38.5&10.3/12.3&31.8/36.9&47.3/54.3&3.7/5.2 \\
\bottomrule
  \end{tabular}
  }
  \label{tab:dual training}
\end{table}
\begin{table}[t]  
\caption{\textbf{Ablation study on individual model components.} MSA indicates the self-attention mask and $cA^{bu}$ refers to the center-biased boxes in BU-Qs initialization.}
  \centering
  \resizebox{0.6\linewidth}{1.2cm}{
  \begin{tabular}{
    @{}c c c c|c c |c c
    @{}
  }
    \toprule
    $S^{sp}$& $S^{ct}$&$MSA$&$cA^{bu}$&{R@50} &{R@100} & {mR@50}& {mR@100}\\
    \midrule
$\checkmark$&$\space$&$\checkmark$&$\checkmark$&33.0&37.5&10.1&12.2\\
$\space$&$\checkmark$&$\checkmark$&$\checkmark$&31.5&36.5&9.5&11.6\\
$\checkmark$&$\checkmark$&$\space$&$\checkmark$&32.8&37.5&9.8&11.7\\
$\checkmark$&$\checkmark$&$\checkmark$&$\space$&32.9&37.8&9.7&12.0\\
$\checkmark$&$\checkmark$&$\checkmark$&$\checkmark$&33.5&38.5&10.3&12.3\\
\bottomrule
  \end{tabular}
  }
  \label{tab:proposed components}
\end{table}

\textbf{TD-Qs and BU-Qs Analysis.} To analyze the effect of TD-Qs and BU-Qs, we compare the results of the Dual-SGG w/o TD-Qs, Dual-SGG w/o BU-Qs, and Dual-SGG. Note that when only a single group of triplet queries is employed, we extend the number of triplet queries to $N_{td}+N_{bu}$ for fair comparison. In addition to R@K and mR@K, we also detail the micro-R@K, micro-UDR@K, and micro-DR@K, as introduced in \cref{sec: intro}, to provide a breakdown analysis from the perspective of detector-conditioned reachability. To this end, we report the performance of the detector-based model introduced in \cref{fig: Detector-based SGG} as the baseline. The sizes of the created Det-T and UDet-T on the VG test set are 98,103 and 54,123. The comparison is shown in \cref{tab:instance-level detectability conditions SGG} above.  Dual-SGG w/o BU-Qs slightly improves SGG performance compared to the baseline by achieving a higher micro-UDR@K. Despite the TD-Qs being designed to approximate the detector-based reasoning mechanism, updating TD-Qs in the context of triplets provides the possibility to explore triplets involving instances that are semantically uncovered by the object detector. However, the micro-UDR@K is still lagging behind Dual-SGG w/o TD-Qs, as TD-Qs struggle to capture the triplets spatially outside the detector-conditioned reachability, as shown in \cref{fig:attn}. Consistent with \cref{tab:Introduction}, the great micro-UDR@K of Dual-SGG w/o TD-Qs comes at the cost of micro-DR@K, which leads to lower overall performance. Dual-SGG achieves the best overall performance and outperforms the baseline and Dual-SGG w/o BU-Qs by increasing micro-UDR@K while maintaining micro-DR@K. This comparison demonstrates the necessity of the co-existence of TD-Qs and BU-Qs under the same total triplet query budget.

\textbf{Strong Detector Condition Case Analysis.}
The performance improvement of Dual-SGG over Dual-SGG w/o BU-Qs primarily arises from its enhanced results on UDet-T, which may be sensitive to the size of UDet-T. If a stronger object detector is employed, reducing UDet-T, the improvement of Dual-SGG may be negligible. To investigate the effect of Dual-SGG under stronger detector conditions, we conduct an additional experiment. Specifically, we train new Dual-SGG, Dual-SGG w/o BU-Qs, and Dual-SGG w/o TD-Qs based on a new object detector constructed by replacing the backbone of the default detector (Deformable DETR) with Swin-B \cite{liu2021swin}, and the entity query ($N_{e}$) is set to 900. The results are shown in \cref{tab:instance-level detectability conditions SGG} below. Although UDet-T shrinks from 35\% to 28\% under a stronger detector condition, our Dual-SGG still remains effective.

\textbf{Complementarity Analysis.} In this section, we provide a detailed comparison to analyze 1) whether Dual-SGG can truly capture the complementary predictive behaviors of specialized detector-based and query-based models, and 2) whether the joint updating of TD-Qs and BU-Qs is necessary. Specifically,  we independently train a Dual-SGG w/o TD-Qs and a Dual-SGG w/o BU-Qs as specialized models, setting the number of triplet queries as $N_{bu}$ and $N_{td}$. The performances of these models are reported in \cref{tab:dual training} above. Subsequently,  $\operatorname{Top}100$ predictions from each specialized model are taken to form a 200-candidate pool. We apply three post-hoc methods, including max, average, and normalization, as alternative fusion strategies that directly calibrate the 200 triplet scores and re-rank them, yielding 100 predictions for the evaluation. As shown in \cref{tab:dual training}, our Dual-SGG outperforms both specialized models on all evaluation metrics, demonstrating that our Dual-SGG is attributable to the complementary predictions of the two specialized models. In addition, the comparison with other fusion strategies reflects the necessity of the joint updating of TD-Qs and BU-Qs.

\textbf{Individual Components.} \cref{tab:proposed components} demonstrates the effect of other individual components in Dual-SGG, including the spatial and content scores in EPS \ie $S^{sp}$ and $S^{ct}$, the self-attention mask in the triplet decoder, and the center-biased initialization of triplet anchors $\{a^{bu}_{i}\}_{i=1}^{N_{bu}}$. The exclusion of $S^{sp}$ or $S^{ct}$ results in degradation, as they provide high-quality positive and negative samples, respectively. When the self-attention mask is removed, the decreased performance demonstrates that the dependency of $R^{bu}$  on $R^{td}$ leads to a negative effect on complementarity. The results in the last two rows illustrate that the center-biased initialization for $\{a^{bu}_{i}\}_{i=1}^{N_{bu}}$ is important for achieving better performance. Note that when center-biased initialization is removed, the anchors of BU-Qs are initialized with uniformly distributed boxes over the image, which is widely adopted in DETR-style models.
\subsection{Qualitative Results}
\begin{figure}[tb]
  \centering
  \includegraphics[height=3.0cm]{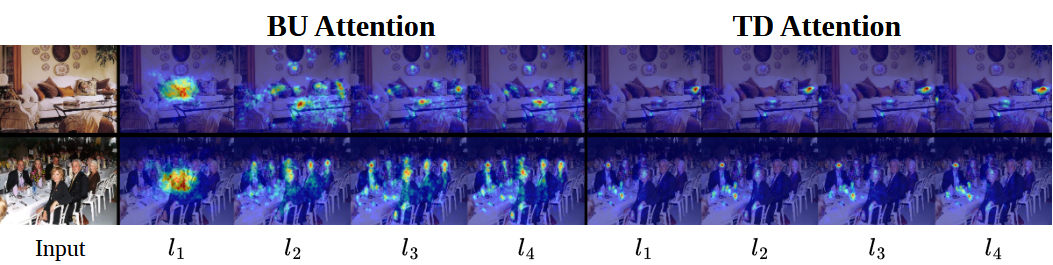}
  \caption{\textbf{Attention visualization.} Columns 2–5 and 6-end depict the attention distributions of BU-Qs and TD-Qs across the decoder layers ($l_{i}$), respectively.}
  \label{fig:attn}
\end{figure}

\begin{figure}[tb]
  \centering
  \begin{subfigure}[b]{0.33\linewidth}
    \centering
    \includegraphics[width=\linewidth]{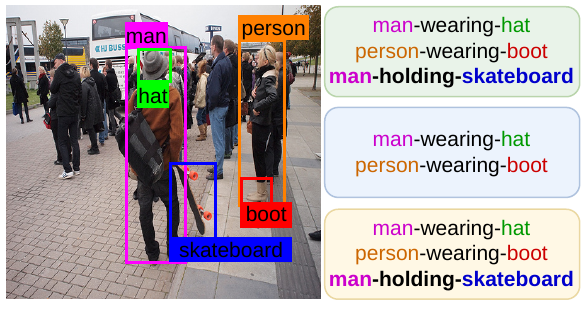}
    \caption{}
    \label{fig:example_vis a}
  \end{subfigure}\hfill
  \begin{subfigure}[b]{0.33\linewidth}
    \centering
    \includegraphics[width=\linewidth]{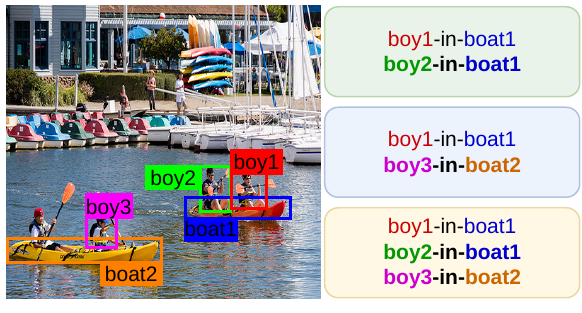}
    \caption{}
    \label{fig:example_vis b}
  \end{subfigure}\hfill
  \begin{subfigure}[b]{0.33\linewidth}
    \centering
    \includegraphics[width=\linewidth]{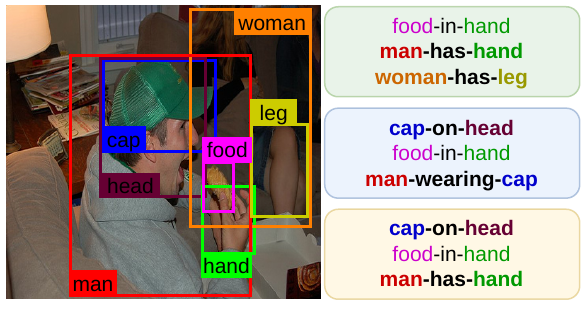}
    \caption{}
    \label{fig:example_vis c}
  \end{subfigure}\hfill
  \label{fig:example_vis}
  \caption{\textbf{Example visualization.} For each image, the detected ground-truth triplets by Dual-SGG w/o TD-Qs (top), Dual-SGG w/o BU-Qs (middle), and Dual-SGG (bottom), are provided. Triplets highlighted in \textbf{bold} denote missed triplets by at least one of the other two models.}
\end{figure}

\textbf{Attention Visualization.} We identify the $\operatorname{Top}100$ triplets from each of $R^{td}$ and $R^{bu}$. The corresponding triplet queries are tracked across the decoder layers. For each layer, the cross-attention maps of the TD-Qs and BU-Qs are aggregated separately to produce heat maps. The heat maps indicate the focused regions of TD-Qs and BU-Qs for exploring triplets. As shown in \cref{fig:attn}, distinct exploration patterns are exhibited. Specifically, BU-Qs initially focus on the center of the image due to the center-biased initialization and gradually spread outwards. In contrast, TD-Qs consistently focus on the regions of the initial entity pairs. This figure illustrates that the BU-Qs can capture the triplets involving instances that are spatially unreachable for the TD-Qs due to the detector constraint. \\
\textbf{Complementarity Visualization.} \cref{fig:example_vis a} and \cref{fig:example_vis b} demonstrate the complementarity property of our Dual-SGG. 
The triplets missed by either Dual-SGG w/o TD-Qs or Dual-SGG w/o BU-Qs can be successfully predicted by Dual-SGG. \cref{fig:example_vis c} illustrates that sometimes the complementarity of Dual-SGG cannot recover every detected ground-truth triplet from Dual-SGG w/o TD-Qs and Dual-SGG w/o BU-Qs. We aim to address this limitation in future work.
\section{Conclusion}
This study investigates the distinct predictive behaviors of detector-based and query-based SGG methods from the perspective of detector-conditioned reachability. The analysis reveals that detector-based methods effectively capture triplets that lie within their detector-conditioned reachability space, whereas query-based methods are capable of identifying triplets beyond this space. Based on these insights, a Dual-SGG method is introduced, achieving superior performance by amalgamating the strengths of the two reasoning mechanisms within a single framework. The end-to-end framework consolidates the distinct predictive behaviors without the need for multiple specialized models and complex fusing strategies.
\section*{Acknowledgements}
 Funded by the Deutsche Forschungsgemeinschaft (DFG, German Research Foundation) under Germany’s Excellence Strategy – EXC 2002/1 “Science of Intelligence” – project number 390523135. This work has been partially supported by the MIAI Cluster - project number ANR-23-IACL-0006, and by the Institute of Information \& Communications Technology Planning \& Evaluation (IITP) grant funded by the Korean Government (MSIT) (No. RS-2024-00457882, National AI Research Lab Project).

%
%
\bibliographystyle{splncs04}
\bibliography{main}
\end{document}